\documentclass[runningheads]{llncs}
\usepackage{color,soul}
\usepackage{xcolor}
\usepackage{graphicx}
\usepackage{textcomp}
\usepackage{caption}
\usepackage{subcaption}
\usepackage{hyperref}
\usepackage{url}
\usepackage{array,multirow,makecell}
\usepackage{breakurl}
\usepackage{mathtools}
\usepackage{tikz}
\usepackage[square,sort,comma,numbers]{natbib}
\usepackage{lipsum}
\usepackage[title]{appendix}

\usepackage{amsmath,amssymb}
\DeclareMathOperator{\E}{\mathbb{E}}

\usepackage{enumitem}
\usepackage{booktabs}
\usepackage{soul}

\definecolor{light-red}{HTML}{FF7E71}
\definecolor{light-blue}{HTML}{64A1FF}
\definecolor{light-yellow}{HTML}{FFCC71}

\newif\ifconsiderlater
\considerlatertrue

\ifconsiderlater
	\newcommand{\todo}[1]{\textcolor{blue}{\textbf{XXX [#1] XXX}}}
\else
\newcommand{\todo}[1]{}
\fi

\begin{document}
\title{Improving Skin Condition Classification with a Visual Symptom Checker Trained using Reinforcement Learning}
%
%
\titlerunning{Improving Skin Condition Classification using Reinforcement Learning}

\author{Mohamed Akrout \inst{1,2}\thanks{Research conducted while working at \href{www.triage.com}{Triage}.}\and
Amir-massoud Farahmand \inst{3,2}\thanks{AMF would like to acknowledge funding from the Canada CIFAR AI Chairs Program.} \and
Tory Jarmain \inst{1} \and \\
Latif Abid \inst{1}}
%
\titlerunning{Improving Skin Condition Classification using Reinforcement Learning}
\authorrunning{M. Akrout et al.}
%
\institute{Triage, Toronto, Canada\thanks{1, Adelaide St. E., Suite 3001, Toronto, M5C1J4, Canada}\\ \and
Department of Computer Science, University of Toronto, Toronto, Canada\\ \and
Vector Institute, Toronto, Canada
}
%
\maketitle              
\begin{abstract}
We present a visual symptom checker that combines a pre-trained Convolutional Neural Network (CNN) with a Reinforcement Learning (RL) agent as a Question Answering (QA) model. This method increases the classification confidence and accuracy of the visual symptom checker, and decreases the average number of questions asked to narrow down the differential diagnosis.
A Deep Q-Network (DQN)-based RL agent learns how to ask the patient about the presence of symptoms in order to maximize the probability of correctly identifying the underlying condition. The RL agent uses the visual information provided by CNN in addition to the answers to the asked questions to guide the QA system. We demonstrate that the RL-based approach increases the accuracy more than 20\% compared  to the CNN-only approach, which only uses the visual information to predict the condition. Moreover, the increased accuracy is up to 10\% compared to the approach that uses the visual information provided by CNN along with a conventional decision tree-based QA system. We finally show that the RL-based approach not only outperforms the decision tree-based approach, but also narrows down the diagnosis faster in terms of the average number of asked questions.

\end{abstract}
\section{Introduction}
Doctors often ask their patients a series of questions in order to narrow down the set of plausible conditions matching the observed symptoms. By asking relevant questions, they can diagnose their patients more efficiently. We are interested in designing an automatic system that diagnoses patients based on an image and a sequence of questions and their corresponding answers.

Given an initial set of symptoms, a QA system such as a symptom checker enables  the  emulation  of  this  conventional  approach by asking the relevant questions in order to refine the differential diagnosis. Some work has been done to formulate the symptom checker using Bayesian networks \cite{zagorecki2013system} and recurrent neural networks \cite{choi2016doctor}. All these symptom checkers, however, rely only on clinical descriptions without leveraging the visual information usually available in several domains such as dermatology and radiology.

The recent work of Akrout et al.~\cite{akrout2018improving} makes use of the visual information by combining a CNN and a decision tree QA model to improve the skin condition classification task. Their proposed decision tree-based approach picks the best symptom to ask by maximizing the information gain $\text{IG}(\mathcal{S}, \mathcal{C})$ between symptoms $\mathcal{S}$ and conditions $\mathcal{C}$. Since decision trees are learned by heuristic methods such as greedy search, they only consider immediate information gain at the current splitting node and often result in sub-optimal solutions in a constrained search space. The RL framework overcomes this problem by searching for splitting strategies in the global search space based on the evaluation of long-term payoff. 

Designing QA systems has already been successfully formulated as an RL task for Query Reformulation \cite{nogueira-cho-2017-task}, search engine querying \cite{chali2015reinforcement}, and automatic diagnosis \cite{wei2018task}.
The latter one is the closest work to ours, in which the authors formulate the problem of learning a QA system for differential diagnoses problems as an RL problem.
A major difference with our work is that they do not use visual information as a prior for the QA system, while this work does, as will be explained in Section~\ref{sec1:symptomchecker-mdp}.
Additionally, they use a different action space, where actions are \textit{inform}, \textit{request}, \textit{deny}, \textit{confirm}, \textit{thanks} and \textit{close\_dialogue}, which is unlike our action space consisting of the presence of the symptoms themselves. Moreover, they use a reward function with hard-coded discrete values based on whether they find the right diagnosis or not, whereas ours is a dense reward given to the agent at each step and is defined based on the probability of choosing the true condition given the sequence of actions taken so far. 
In this work, we extend the study in \cite{akrout2018improving} by formulating the symptom checking problem as a Markov Decision Process (MDP). We show that the RL agent learns using DQN to ask the best symptom and outperforms the decision tree approach while asking fewer questions. This paper makes the following contributions:
\begin{itemize}
    \item We formulate the visual symptom checking problem as an MDP problem, and propose an RL-based approach to solve it.
    \item We show a significant accuracy improvement compared with the decision tree approach.
    \item We illustrate how the RL approach not only significantly improves the accuracy of the skin disease classification compared to the decision tree approach but also decreases the average number of asked questions.
\end{itemize}



\section{A Visual Symptom Checker as an MDP}\label{sec1:symptomchecker-mdp}
A visual symptom checker interacts with the patient by asking  symptoms and receiving the patient's answer. In this section, we show how the visual symptom checking problem can be formulated as an MDP \cite{puterman1994markov,sutton2018reinforcement}.
\subsection{MDP Definition}\label{sec:mdp-definition}
An MDP is a tuple ($\mathcal{X}$, $\mathcal{A}$, $\mathcal{P}$, $\mathcal{R}$, $\gamma$) where $\mathcal{X}$ is the set of states, $\mathcal{A}$ is a finite set of actions from which the agent can choose, $\mathcal{P}$ : $\mathcal{X}$ $\times$ $\mathcal{A}$ $\times$ $\mathcal{X}$ $\rightarrow$ [0, 1] is a transition probability in which $\mathcal{P}(x, a, x')$ defines the probability of observing state $x'$ after executing action $a$ in the state $x$, $\mathcal{R}: \mathcal{X} \times \mathcal{A} \rightarrow \mathbb{R}$ is the expected reward after being in state $x$ and taking action $a$, and $\gamma$ $\in$ [0, 1) is the discount factor.

An RL agent continually makes value judgments in order to select the right action. The action selection mechanism of an RL agent is called its policy, which is a mapping $\pi: \mathcal{X} \rightarrow \mathcal{A}$ from the state space to the action space. Given a policy $\pi$, the action-value function $Q^\pi: \mathcal{X} \times {A} \rightarrow \mathbb{R}$ is
\begin{equation}
\label{eq:state-action-function}
Q^\pi(x,a) = \E\bigg[\; \sum_{k=0}^{\infty} \gamma^k r_{t+k+1} \;|\; X_t=x, A_t=a\;\bigg].
\end{equation}
The action-value function $Q^\pi$ at state-action pair $(x,a)$ is the expected discounted reward that the RL agent receives if it starts from state $x$ at time $t$, chooses action $a$, and afterwards selects actions according to policy $\pi$, i.e., $A_{t+k} = \pi(X_{t+k})$ for $k =1, 2, \dotsc$. The goal of an RL agent is to find a policy $\pi$ that maximizes this value for all state-action pairs. Such a policy is called the optimal policy $\pi^*$, and its corresponding action-value function is called the optimal action-value function $Q^*$.
If the agent has access to the optimal action-value function, the optimal policy $\pi^*$ can be computed as $\pi^*(x) \leftarrow \arg \max_{a \in \mathcal{A}} Q^*(x,a)$. Many RL algorithms, including Deep Q-Network (DQN)~\cite{mnih2015human} that we briefly describe in Section~\ref{sec:dqn}, try to estimate $Q^*$. For more information, refer to~\cite{sutton2018reinforcement}.


We show how the MDP framework can be applied to the symptom checking problem in two steps: we first describe in Section \ref{sec:mdp-formulation} how $\mathcal{X}$, $\mathcal{A}$, $\mathcal{P}$ and $\mathcal{R}$ are designed for the symptom checker. Afterwards, we describe in Section \ref{sec:dqn} the agent's algorithm that improves the skin condition classification.
\subsection{MDP Formulation}\label{sec:mdp-formulation} 
Let $\mathcal{S}$ be the symptom space. Some examples are rash, redness and pigmented lesion. Let $\mathcal{C}$ be the condition space. Some examples are cellulitis, psoriasis and melanoma, to name a few. We denote any symptom by $s_i$ and any condition by $c_j$ where $1\leq i\leq|\mathcal{S}|$, $1\leq j \leq |\mathcal{C}|$, and $|\cdot|$ refers to the cardinality operator. Let $\mathcal{S}^a \subset \mathcal{S}$ be the set of asked symptoms and $s^a$ any asked symptom among $\mathcal{S}^a$.

\begin{figure}[h!]
\centering
\includegraphics[width=0.6\linewidth]{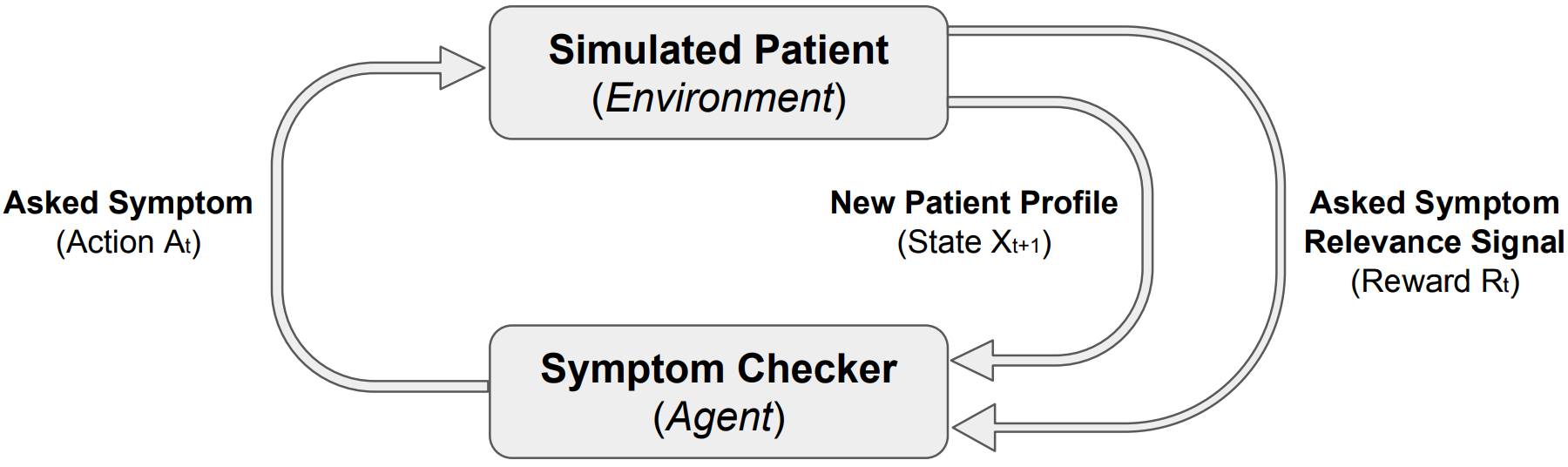}
\caption{The agent-patient interaction in an MDP.}
\label{fig:MDP}
\end{figure}
A symptom checker agent learns which symptom $s \in \mathcal{S}$ to ask for an environment of a simulated patient having a condition $c^* \in \mathcal{C}$. Figure~\ref{fig:MDP} summarizes pictorially how a symptom checker can be viewed as an RL agent finding an optimal policy for an MDP.
We define the MDP parameters as follows:
\begin{itemize}
    \item[$\bullet$] \textbf{Action space $\mathcal{A}$}: It corresponds to the symptom space $\mathcal{S}$ where each action $a \in \mathcal{A}$ is a possible symptom $s$ to ask.
    \item[$\bullet$] \textbf{State space $\mathcal{X}$}: Each state $x \in \mathcal{X}$ corresponds to a patient state. A state $x$ is the concatenation of the patient's history of all answers and the pretrained CNN's output probabilities of the patient's image (see Fig~\ref{fig:patient-env}).
    \begin{figure}[t]
        \centering
        \includegraphics[width=0.6\linewidth]{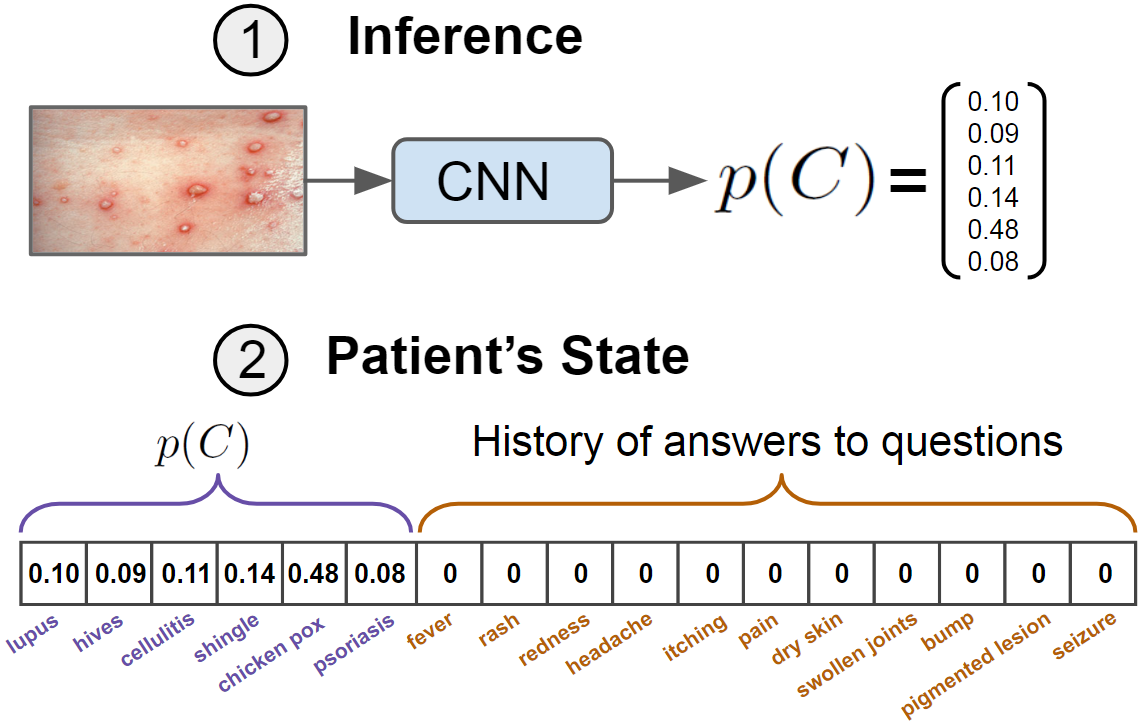}
        \caption{State design of the simulated patient's environment: (1) CNN inference on the patient's image to compute $p(C)$ which is a vector of $|\mathcal{C}|$ elements. Here $|\mathcal{C}|=6$, (2) The state of the simulated patient is a concatenation of $p(C)$ with the history of answers to questions, which is a vector of $|\mathcal{S}|$ elements. Here $|\mathcal{S}|=11$.}
        \label{fig:patient-env}
    \end{figure}
    \item[$\bullet$] \textbf{Transition probability $\mathcal{P}$}: For an asked question, +1 or -1 are put at the question's position if the answer is respectively ``yes'' or ``no'' (see Fig~\ref{fig:state-patient-env}).
 \begin{figure}[t]
  \begin{subfigure}[b]{0.5\textwidth}
    \includegraphics[width=1\textwidth]{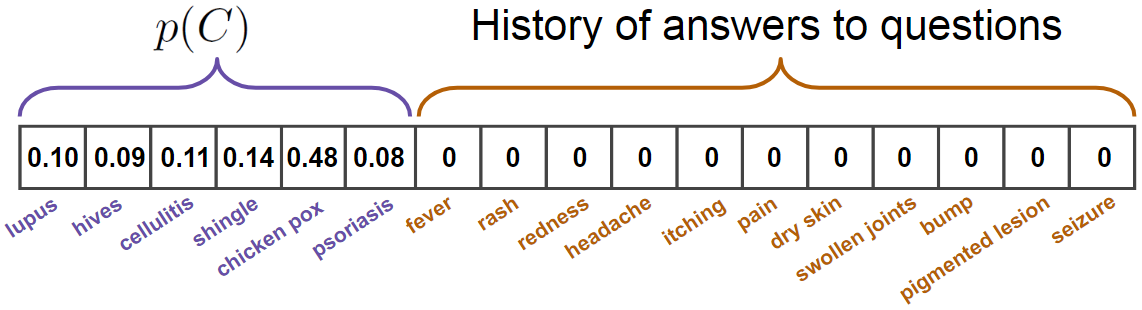}
    \caption{The initial environment state}
    \label{fig:1}
  \end{subfigure}
  \begin{subfigure}[b]{0.5\textwidth}
    \includegraphics[width=1\textwidth]{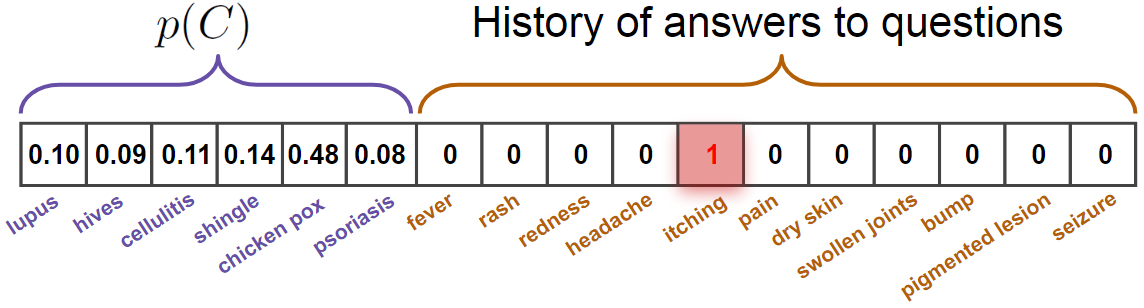}
    \caption{The updated environment state}
    \label{fig:2}
  \end{subfigure}
  \caption{An example of the update of environment state: (a) the vector of history answers is set to 0, (b) the environment state gets updated by putting +1 in the ``itching'' position after answering ``yes'' to the asked symptom ``itching''.}
  \label{fig:state-patient-env}
\end{figure}
    For each new random patient simulation, its condition $c_i=c^*$ is known. The simulated patient answers the asked symptom $s^{a}$ by referring to the health knowledge matrix $\mathcal{M}$ \cite{rotmensch2017learning} computed from electronic medical records of 273,174 patients. It represents a matrix of condition-symptom relationships where each cell (i, j) represents the conditional probability $p(s_j | c_i)$ of a symptom $s_j$ given a condition $c_i$. The simulated environment answers $s^{a}$ by generating a random number $k$ from a uniform distribution between 0 and 1. If $k \leq p(s^{a} | c^*)$, the answer is ``yes'', and ``no'' otherwise.\\
    \item[$\bullet$] \textbf{Reward $\mathcal{R}$}: This function is equal to $p(c^*|S^a)$, the probability of correct condition $c^*$ given the asked questions $S^a$. This design of the reward function is mainly based on the intuition that we want to keep the probability of the true condition of the simulated patient as high as possible. At each interaction, when the simulated patient answers $s^a$, we update the conditional probability $p(c_i|S^a)$ $\forall i$ using the Bayes rule as described in the Appendix~\ref{App:A}.
\end{itemize}


\subsection{Deep Q-Learning}\label{sec:dqn}

The visual symptom checker agent learns an estimate of the optimal action-value function $Q^*$. Mnih et al. \cite{mnih2015human} introduced the Deep Q-Network (DQN), a CNN approximating the action-value function $Q(x,a; \omega) \approx Q^*(x,a)$, where $\omega$ represents
the network's parameters. The DQN squared error loss function $L(\omega)$ is defined as:
\begin{equation}
\label{eq:dqn}
L(\omega) = \E\bigg[\,\bigg( r + \gamma \max_{a'}\;Q_{\text{target}}(x', a'; \omega^{-}) - Q_{\text{net}}(x, a; \omega)\bigg)^2\,\bigg],
\end{equation}
where $x$ is the current state, $x'$ is the next-state, and $\gamma$ is the discount factor.
DQN uses $Q_{\text{target}}(\omega^{-})$, a fixed version of $Q_{\text{net}}(\omega)$ with parameters $\omega^{-}$ that are periodically updated in order to stabilize rapid policy changes, due to the quick variations in Q-values. Another trick used in \cite{mnih2015human} to avoid divergence because of successive data sampling is the experience replay buffer. It stores transitions of ($x$, $a$, $r$, $x'$) and is randomly sampled to create the mini-batches used for training.

\section{Classification of Skin Conditions}\label{sec:classification}
In this section, we describe both the training and inference setting of the visual symptom checker. In our setting, we consider $|\mathcal{S}|=330$, $|\mathcal{C}|=9$, and the discount factor $\gamma=0.99$.
\subsection{The Training Step}
We initially train an Inception-v3 network \cite{szegedy2016rethinking} on a dataset consisting of 5,841 images that are equally divided into 9 skin conditions: atopic dermatitis, lupus, shingles, cellulitis, chickenpox, hives, psoriasis, gout and melanoma. We have used a data split of 70\%-15\%-15\% for training, validation and testing the CNN, respectively. 
For a given patient image, we run inference on the pretrained CNN to get the CNN's output probabilities which, concatenated with the history of answers initialized at zero, form the environment's state of the simulated patient. The RL agent interacts with this environment until either a predefined maximum number of questions to ask is reached or one of the 9 condition probabilities $p(c_i|S^a)$ exceeds a predefined threshold.

\subsection{The Inference Step}
We run inference on our visual symptom checker using annotated test images (i.e. a labeled image with a paragraph describing the patient's symptoms). The image description, commonly called a \textit{vignette}, does not include the true condition of the patient, but rather their symptoms. We answer the asked questions of the trained RL agent with ``yes'' if the symptom is present in the vignette and with ``no'' otherwise. The Appendix \ref{App:appendix-vignettes} provides 9 examples of vignettes, one for each condition.

\section{Experiments and Results}\label{sec:results}
All the experiments have been performed within the simulated patient environment described in Section \ref{sec:mdp-formulation}. We fix the maximum number of questions to ask at 10 and the confidence threshold to 95\% respectively.
\subsection{Architecture of the Q-network (DQN)}
The Q network has 5 layers: an input layer of 339 units (a vector of $|\mathcal{S}|=330$ elements representing the history of answers concatenated with a vector of $|\mathcal{C}|=9$ probabilities from the CNN output), 3 hidden fully connected layers consisting of 350 ReLu units and an output layer composed of $|\mathcal{S}|$ = 330 linear units.
\subsection{Evaluation}
We have evaluated our visual symptom checker using 600 annotated test images. The evaluation metric chosen to compare the system before and after the visual symptom checker is the top-K accuracy, which is well known in medical imaging scenarios to successfully assess the differential diagnosis cases. Additionally, we compare the average number of asked questions between the RL approach and the decision tree approach in \cite{akrout2018improving}. 
\subsubsection{Top-K accuracy}
Table~\ref{results} shows how the RL agent that learns from a simulated patient environment increases the rank and the probability of the correct condition compared to the classification performance of the CNN alone.
\setlength{\tabcolsep}{0.6em} 
\begin{table}[tb]
\centering
\small\addtolength{\tabcolsep}{4pt}
\caption{The CNN individual and combined classification performance with two QA models: RL and decision tree.}
\label{results}
\renewcommand{\arraystretch}{1}
\begin{tabular}{|c|c|c|c|} 
       \cline{2-4}
       \multicolumn{1}{c|}{} & \multirow{2}{*}{\textit{CNN model}} &  \multicolumn{2}{c|}{\textit{CNN + QA model}} \\ \cline{3-4} 
       \multicolumn{1}{c|}{} & & Decision Tree  & RL  \\\hline
       \multicolumn{1}{|c|}{Top-1} & 49.75 $\pm$ 0.2\% &  57.64 $\pm$ 0.3\%  & \textbf{70.41 $\pm$ 0.2\%}  \\\hline
       \multicolumn{1}{|c|}{Top-2} & 63.55 $\pm$ 0.6\% &  73.01 $\pm$ 0.5\%  & \textbf{82.45 $\pm$ 0.1\%}   \\\hline
       \multicolumn{1}{|c|}{Top-3} & 80.29 $\pm$ 0.5\% &  85.62 $\pm$ 0.5\% & \textbf{91.22 $\pm$ 0.2\%}  \\\hline
 \end{tabular}
\end{table}
Here we show five-fold cross-validation classification accuracy for the CNN model. In each fold, a different fifth of the dataset is used for validation, with the rest of the dataset used for training. For the CNN model, reported values are the mean and standard deviation of the validation accuracy across all $n=5$ folds.

The results of the decision tree QA model are those of \cite{akrout2018improving}. For the RL approach, 5 independent agents were trained on one of the 5 CNNs corresponding to a specific cross-validation fold. Reported values are the mean and standard deviation of the evaluation accuracy across all the 5 environments.
These results demonstrate that the proposed RL approach can significantly improve the classification performance across a multiclass classification task and outperform the decision tree approach used in \cite{akrout2018improving}.

By sampling many times, the RL agent asymptotically learns, as the number of simulated patients increases, the conditional probabilities between symptoms and conditions. For instance, given two specific symptom $s_0$ and condition $c_0$, if $p(s_0|c_0) = 60\%$, the environment will answer 60\% of the times ``yes''. Therefore, 60\% of the data collected by the agent from a simulated patient with a condition $c_0$ will have $s_0$ included. The fact that the training data is balanced proportionally to the symptoms' occurrences allows the agent to learn the probabilistic relationships which is one reason explaining why the RL approach can outperform greedy approaches like decision trees. Another reason is that RL sequentially backtracks rewards unlike decision trees which search through the space of possible branches with no backtracking.

\subsubsection{Average Number of Asked Questions}
\begin{figure}[tb]
\centering
\includegraphics[width=0.4\linewidth]{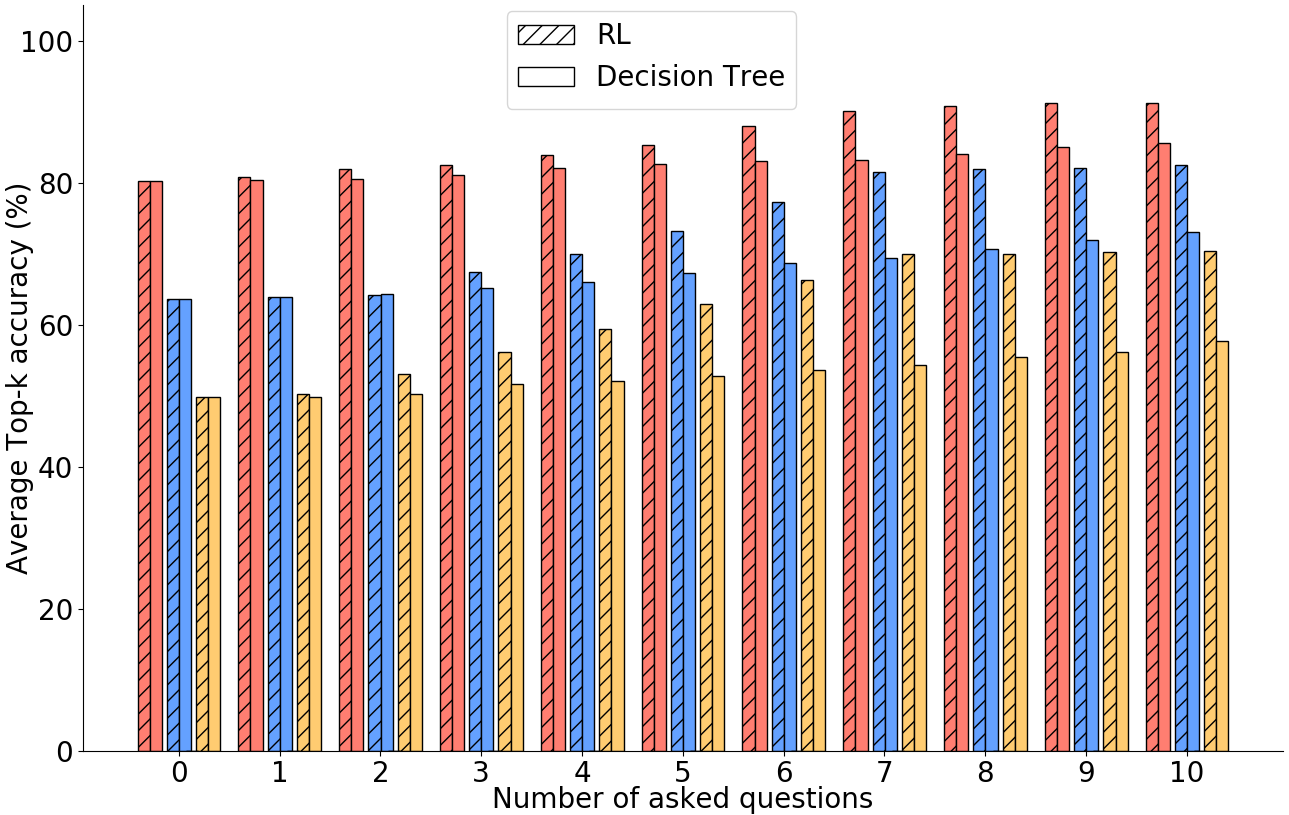}
\caption{Evolution of the top-3 (\protect\tikz \fill[light-red] (0.1,0.0) rectangle (0.4,0.2);), top-2 (\protect\tikz \fill[light-blue] (0.1,0.0) rectangle (0.4,0.2);) and top-1 (\protect\tikz \fill[light-yellow] (0.1,0.0) rectangle (0.4,0.2);) accuracy with the number of asked questions. The top-K accuracy with a decision tree as QA model keeps increasing slowly. The top-K accuracy rate of the RL agent as a QA model is not only higher than the decision tree method but also converges quicker in around 7 questions.}
\label{fig:nquestions}
\end{figure}

While the reported numbers of Table~\ref{results} are obtained after a maximum number of 10 questions, we examined the evolution of the top-K accuracy after answering each asked question by both the proposed RL-based QA model and the decision tree model studied in \cite{akrout2018improving} on the same dataset. Figure~\ref{fig:nquestions} shows the evolution of the top-K accuracy with the number of asked questions for both approaches. One can see that both top-K accuracy and top-K rate of the RL approach are higher compared to the decision tree approach.

Additionally, the RL agent achieves its highest performance with fewer questions compared to the decision tree approach where top-K accuracy keeps increasing with a lower top-K accuracy rate as the number of questions increases.
This suggests that the RL agent succeeds not only to narrow down the differential diagnosis with a better performance, but also quicker than the decision tree approach.

\section{Conclusion}
\label{sec:conclusions}
The results of the visual symptom checker formulated as an RL problem demonstrate that the proposed methodology can effectively not only learn the probabilistic relationships between symptoms and conditions, but also find the most relevant questions to ask for improving the final predictions across the nine skin conditions considered. Future work remains as to extend the symptom checker to support more conditions and symptoms and to further look into ways to evaluate its performance against doctors and other diagnostic systems.

\subsubsection*{Acknowledgements} We thank the rest of the Triage team for assisting with infrastructure and evaluation, as well as with providing feedback and helpful discussions. We acknowledge Mitacs Accelerate program for the funding they provide.

\let\oldbibliography\bibliography
\renewcommand{\bibliography}[1]{{%
  \let\chapter\section
  \oldbibliography{#1}}}

{
\bibliographystyle{splncs04}
\bibliography{micaai2019.bib}
}
\newpage
\begin{subappendices}
\renewcommand{\thesection}{\Alph{section}}
\section{Details of updating the posterior $p(c_i|S^a)$ $\forall i$}\label{App:A}
Using the vector notation, we denote $p(c_i)$ $\forall i$ by $p(C)$. As described in Section \ref{sec:mdp-formulation}, the simulated patient uses a matrix $\mathcal{M}$ representing the conditional condition-symptom relationships $p(S | C)$.\\
Given the initial $p(C)$ provided by the CNN's output probabilities, the updated $p(C)$ after asking a set of symptoms $S^a$ is $p(C|S^a)$ that can be computed using the Bayes rule:
\begin{equation}
\label{eq:bayes-rule}
p(C|S^a) \propto p(S^a|C) \cdot p(C)\\
\end{equation}

The likelihood $p(S^a|C)$ is a matrix computed using on both the $\mathcal{M}$ representing $p(S | C)$ and the simulated patient's answers. If the simulated agents answers ``yes'' to the asked symptom $s^a$, we append the column $p(s^a|C)$ to $p(S^a|C)$. If the answer is ``no'', we append $1-p(s^a|C)$ to $p(S^a|C)$.

\section{Examples of vignettes}\label{App:appendix-vignettes}


The dataset used in this study is collected by Triage Technologies Inc., and consists of 5,841 images of 9 equally divided skin conditions: atopic dermatitis, lupus, shingles, cellulitis, chickenpox, hives, psoriasis, gout and melanoma.
The dataset includes samples from all six Fitzpatrick skin types.
The ground truth condition for each image is determined by expert dermatologists. We share below vignettes of the nine supported conditions.


\begin{table}[]
\centering
\scriptsize
\renewcommand{\arraystretch}{1.5}
\begin{tabular}{|c|p{7cm}|p{4cm}|}
\hline
 \textbf{Condition} & \multicolumn{1}{c|}{\textbf{Vignette}} & \multicolumn{1}{c|}{\textbf{Image}} \\ \hline
 Chicken pox &  A 3-year-old boy is noted by his parents to have two small red papules on his abdomen just before bedtime. The next morning he feels hot and has developed multiple lesions over his trunk, some of which look like blisters. Later that day he develops vomiting and diarrhoea and his parents take him to the accident and emergency department. The child has previously been well and attends nursery four days per week. No other family members are obviously affected. Examination The child feels hot and looks slightly miserable, he does not obviously look dehydrated. There are multiple red papules and vesicles over the child’s abdomen, back and around his neck. Most of the vesicles are intact, however some have ruptured and are starting to crust. Mucous membranes are normal. &  \raisebox{-\totalheight}{\includegraphics[width=1\linewidth]{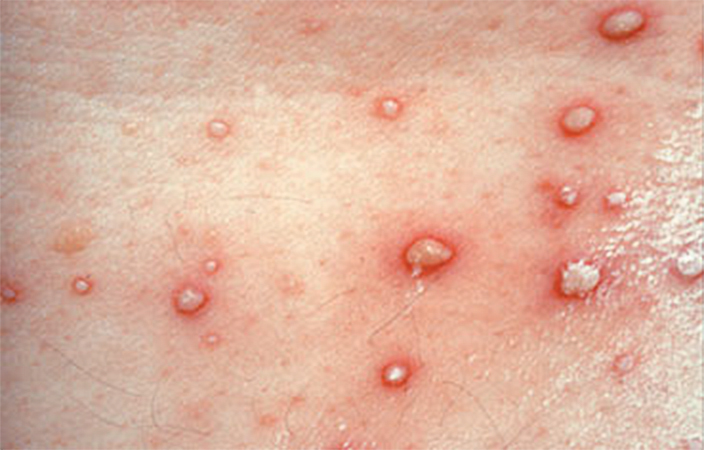}}\\ \hline
 Shingle &  A 29-year-old male with no past medical history presented with bumps on his forehead, behind his left ear, and near his left eye. One week prior, the patient was accidentally hit on his left forehead by a car door while getting in. The next day, he woke up with a bump on his left frontal forehead area where he had been hit. The following day, he noticed a bump behind his left ear. Two days after that, his left medial eye area started to become red and then developed bumps. While his forehead was itchy, none of the other bumps or associated areas were painful. He was afebrile and felt well. He reported having chickenpox as a child. He had tried moisturizer and ice on the lesions; however, neither helped. &  \raisebox{-\totalheight}{\includegraphics[width=1\linewidth]{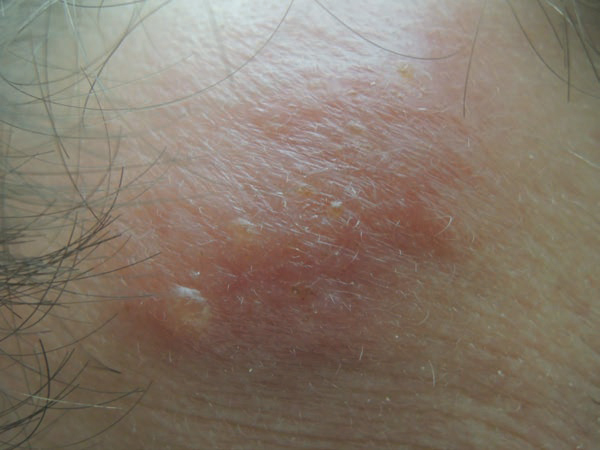}}\\ \hline

\end{tabular}
\end{table}
\newpage

\begin{table}[]
\centering
\scriptsize
\renewcommand{\arraystretch}{1.5}
\begin{tabular}{|c|p{7cm}|p{4cm}|}

\hline
  Atopic-dermatitis & An 11-year-old girl presents to the pediatric dermatology clinic as a referral for a persistently itchy rash that involves the face, neck, trunk and extremities. Her mother states that the girl scratches rigorously at night and is unable to concentrate during school due to pruritus. She has a past medical history of moderate intermittent asthma and allergic rhinitis, and her mother reports a strong history of asthma on her side of the family without any dermatologic issues.  &  \raisebox{-\totalheight}{\includegraphics[width=1\linewidth]{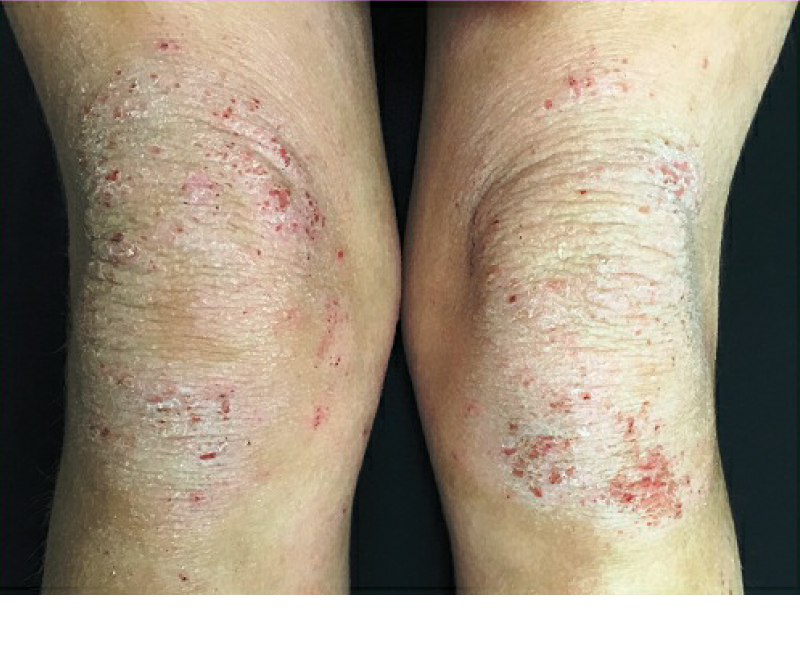}}\\ \hline
Lupus &  I have an odd little situation that has been going on for a couple years now. It gets markedly worse during the winter. I have this chronically purpuric color to my toes year round. It's gotten worse over the past two years. During the winter, I have intermittent episodes (about 3-4/day). My toes get super swollen, red, a purple brown color, and very itchy (it feels like I'm in an ant bed). Episodes last about 30-45 min if I don't itch them or exposed to heat (hot bath, exercise, shoes). &  \raisebox{-\totalheight}{\includegraphics[width=1\linewidth]{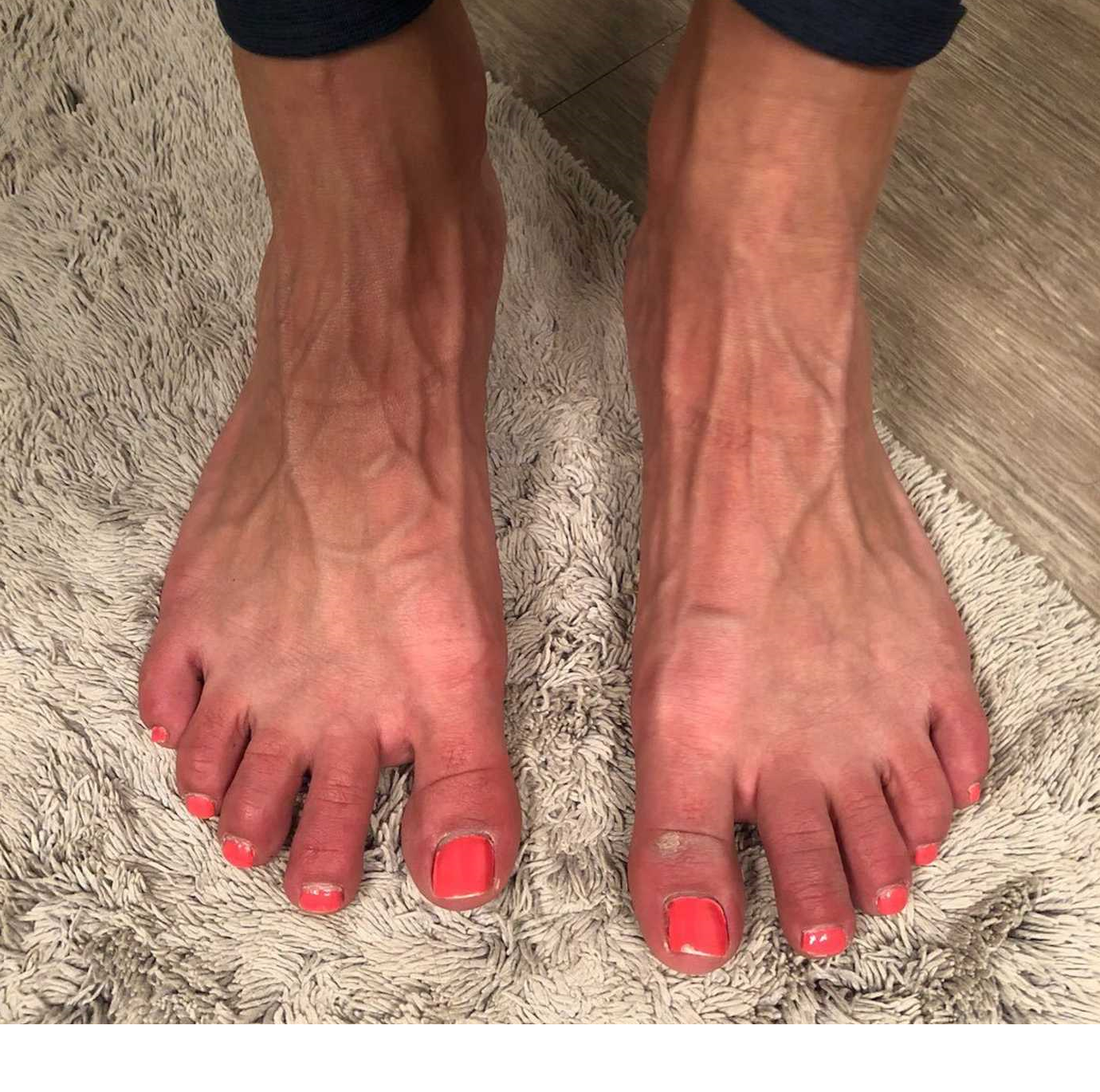}}\\ \hline
Psoriasis &  A 47-year-old man with a pruritic scaly eruption on his body. He states that he began developing similar lesions in his 30s, but over the last several years developed more widespread disease. He notes that his disease is quiescent over the summer months and tends to flare during the winter. He has tried over-the-counter anti-itch creams and other moisturizers without success. His father had a similar skin disease. &  \raisebox{-\totalheight}{\includegraphics[width=1\linewidth]{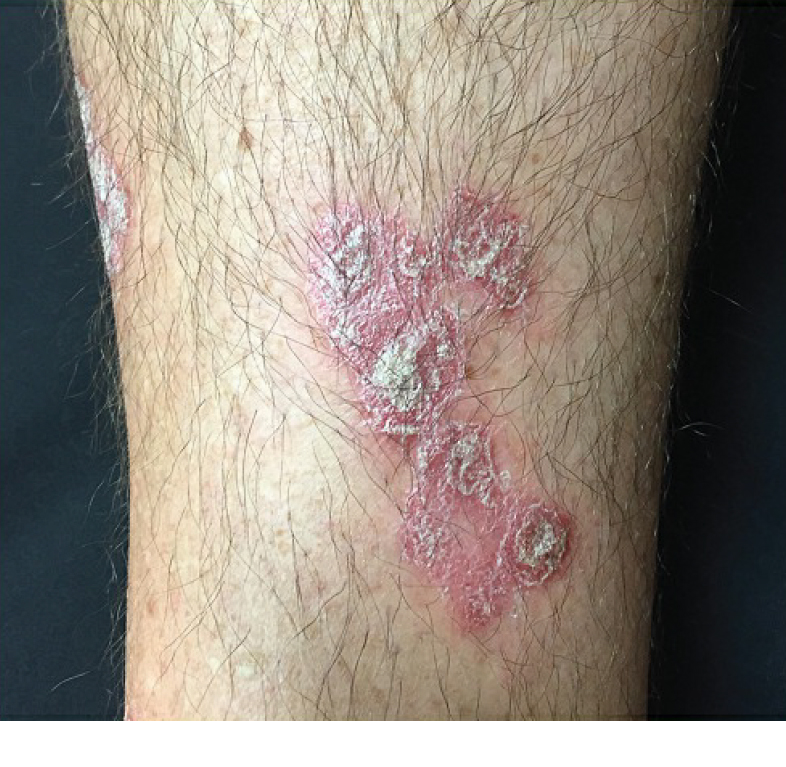}}\\ \hline
 Gout & A 55-year old male with severe ear inflammation, swelling, and multiple bumps on his right ear. He also complains of severe pain, swelling, and tenderness in both hands. On examination, he has a temperature of 100.1\textdegree{}F (37.8\textdegree{}C). There is diffuse warmth, mild erythema, and pitting edema over the dorsum of both hands. There is tenderness and limited hand grip bilaterally. There are multiple uric acid tophi located on the helix of his right ear.&  \raisebox{-\totalheight}{\includegraphics[width=0.7\linewidth]{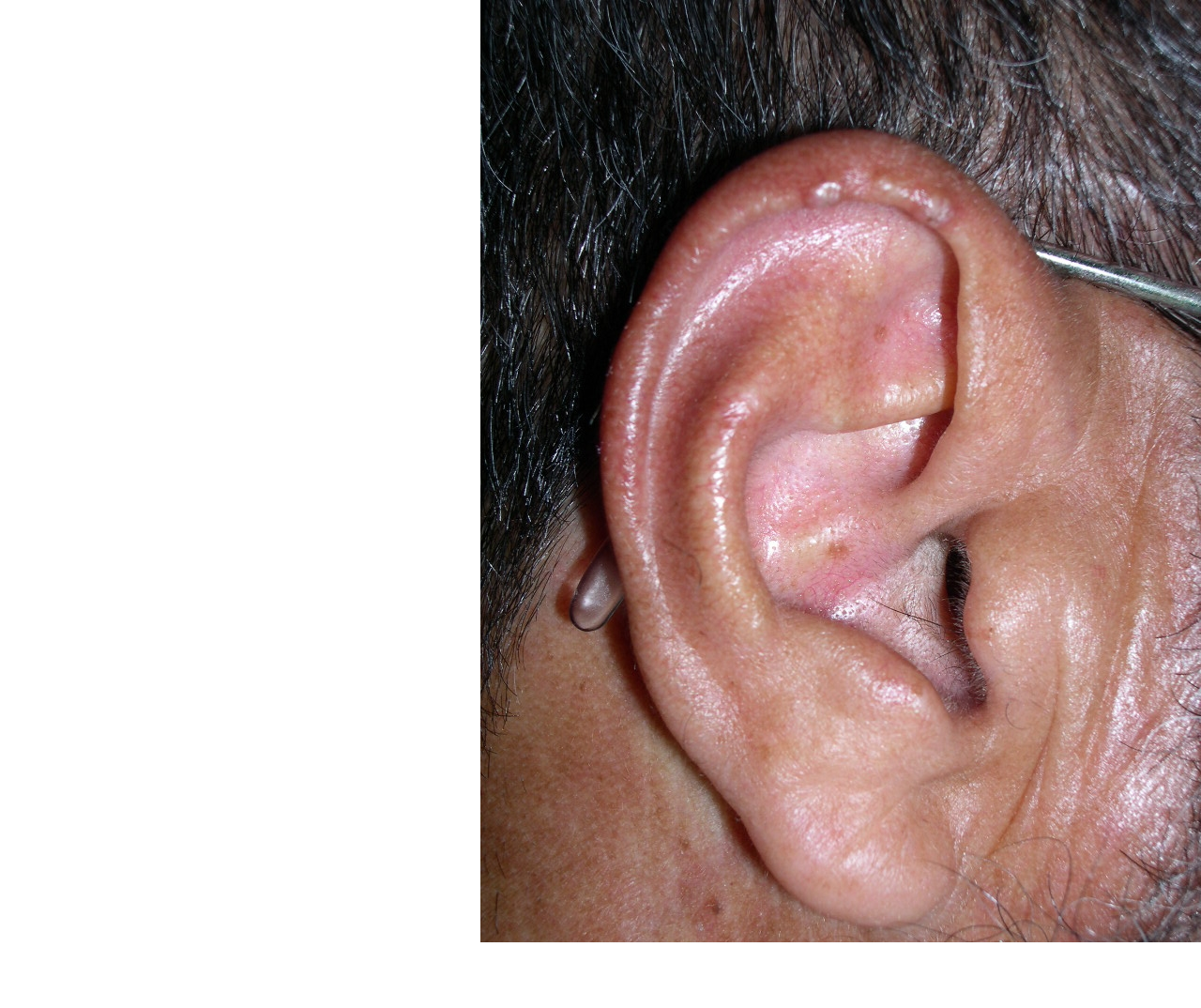}}\\ \hline

\end{tabular}
\end{table}

\begin{table}[]
\centering
\scriptsize
\renewcommand{\arraystretch}{1.5}
\begin{tabular}{|c|p{7cm}|p{4cm}|}
\hline
 \textbf{Condition} & \multicolumn{1}{c|}{\textbf{Vignette}} & \multicolumn{1}{c|}{\textbf{Image}} \\ \hline
  Cellulitis &  A 24-year-old male with past medical history of staphylococcal and streptococcal skin infections and abscesses presented with an open lesion on his right posterior ankle accompanied by right leg erythema and edema. The patient reported that he had formed a blister in this same right heel 7 months prior. That blister healed, however the overlying skin did not fully return to normal. The patient said, “there was always a dark spot there after that.” Five days prior to this visit, his right foot and leg started swelling and becoming red, and the site of the prior blister opened up. The day prior to this visit, a “black spot” appeared adjacent to the open lesion, which then also “broke open,” forming a large wound. He denied fever or chills, and felt otherwise well. The patient reported that the area had been much more painful 2–3 days prior to this visit, at which time he could not put on a shoe due to the pain and swelling. The pain had improved by the time of visit. Vitals were normal. On exam, there was a 3.5 cm × 2 cm weeping ulcer on the lateral Achilles area of his right heel, with surrounding blanchable erythema on his foot and ankle that extended to the mid lower leg. There was 1+ edema on the right lower leg. &  \raisebox{-\totalheight}{\includegraphics[width=1\linewidth]{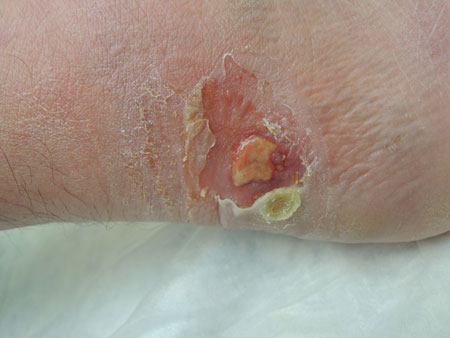}}\\ \hline
   Hives &  A 29-year-old female with no past medical history presented with a diffuse and itchy rash. The rash was most severe on her anterior and posterior lower extremities bilaterally, and less prominent on her lower abdomen and lower forearms. The rash started 2 days prior to her visit and got progressively worse over that time. She felt well otherwise. She reported no new lotions, detergents, or clothes. She got a flu shot 2 days prior (which she had received in the past without incident and she denied an egg allergy). The rash started later in the afternoon after she got her flu shot. The rash on her hands worsened with contact with hot water. The day before and the day of the visit, and the rash continued to progress and to become more itchy. &  \raisebox{-\totalheight}{\includegraphics[width=1\linewidth]{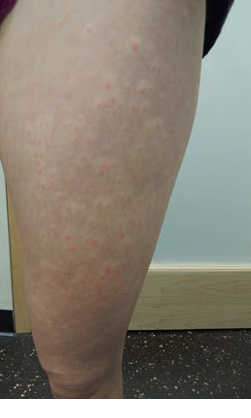}}\\ \hline
    Melanoma &  A 27-year-old woman with a pigmented skin lesion on her left anterior shoulder. She was originally seen four months earlier, and the lesion was noted to be lonely. She had been given a follow-up appointment to be reviewed in six months. She made the decision to present earlier because the lesion had changed in color. This change is concerning for her. &  \raisebox{-\totalheight}{\includegraphics[width=1\linewidth]{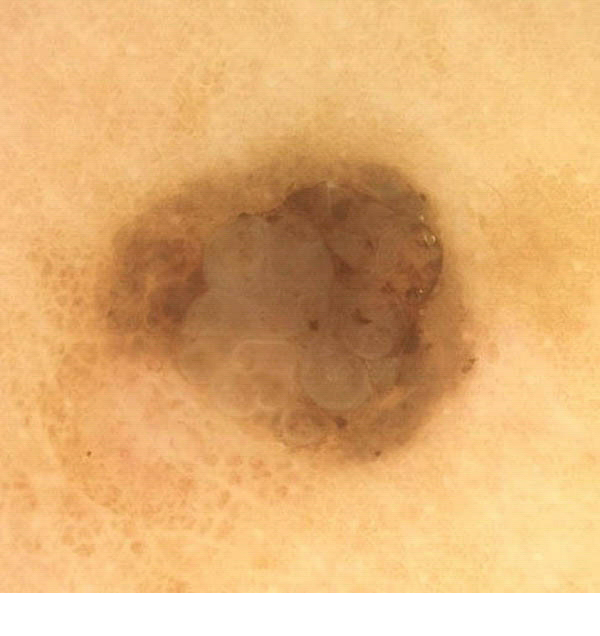}}\\ \hline
\end{tabular}
\end{table}

\end{subappendices}
\end{document}